\documentclass[10pt,twocolumn,letterpaper]{article}

\usepackage{cvpr}
\usepackage{times}
\usepackage{epsfig}
\usepackage{graphicx}
\usepackage{amsmath}
\usepackage{amssymb}
\usepackage{mathtools}
\usepackage{textcomp}
\usepackage{caption}
\usepackage{subcaption}
\usepackage{mathrsfs}
\usepackage{caption}
\usepackage{subcaption}
\usepackage[pagebackref=true,breaklinks=true,letterpaper=true,colorlinks,bookmarks=false]{hyperref}

\cvprfinalcopy 

\def\cvprPaperID{5288} 

\ifcvprfinal\fi
\begin{document}

\title{The Angel is in the Priors:  Improving GAN based Image and Sequence Inpainting with Better Noise and Structural Priors
}

\author{Avisek Lahiri$^{\dagger}$$^{*}$$^1$, Arnav Jain$^{\dagger}$$^2$, Prabir Kumar Biswas$^1$\\$^1$Dept. of E\&ECE, Indian Institute of Technology, Kharagpur\\$^2$Dept. of Mathematics \& Computing, Indian Institute of Technology Kharagpur\\$^{\dagger}$ Equal Contribution\\$^{*}$Email: \href{mailto:avisek@ece.iitkgp.ac.in}{avisek@ece.iitkgp.ac.in} }

\maketitle

\begin{abstract}
Contemporary deep learning based inpainting algorithms are mainly based on a hybrid dual stage training policy of supervised reconstruction loss followed by an unsupervised adversarial critic loss. However, there is a dearth of literature for a fully unsupervised GAN based inpainting framework. The primary aversion towards the latter genre is due to its prohibitively slow iterative optimization requirement during inference to find a matching noise prior for a masked image. In this paper, we show that priors matter in GAN: we learn a data driven parametric network to predict a matching prior for a given image. This converts an iterative paradigm to a single feed forward inference pipeline with a massive 1500$\times$ speedup and simultaneous improvement in reconstruction quality. We show that an additional structural prior imposed on GAN model results in higher fidelity outputs. To extend our model for sequence inpainting, we propose a recurrent net based grouped noise prior learning. To our knowledge, this is the first demonstration of an unsupervised GAN based sequence inpainting. A further improvement in sequence inpainting is achieved with an additional subsequence consistency loss. These contributions improve the spatio-temporal characteristics of reconstructed sequences. Extensive experiments conducted on SVHN, Standford Cars, CelebA and CelebA-HQ image datasets, synthetic sequences and ViDTIMIT video datasets reveal that we consistently improve upon previous unsupervised baseline and also achieve comparable performances(sometimes also better) to hybrid benchmarks.
\end{abstract}
\section{Introduction}
Image inpainting usually refers to filling up of holes or masked regions with plausible pixel values coherent with the neighborhood context. Traditional techniques \cite{barnes2009patchmatch,hays2007scene} were mainly successful in inpainting background and scenes with repetitive textures by matching and copying background patches into holes. However, these methods fail on cases where patterns are unique or non repetitive such as on faces and objects. Also, these methods fail to capture higher semantics of the scene. With the recent breakthrough in generative models such as Variational Autoencoeder (VAE)\cite{kingma2013auto} and Generative Adversarial Networks (GAN) \cite{goodfellow2014generative}, inpainting, in general, is seen as an image completion problem. There are mainly two schools of approach, viz. a) completely unsupervised: conditioned on a prior latent/noise vector \cite{yeh2017semantic} b) mixture of supervised + unsupervised: conditioned on masked image \cite{context_encoders,iizuka2017globally}. The latter methods heavily depend on an initial phase of fully supervised training (reconstruction loss between original and inpainted outputs within the mask), followed by refinement stage with adversarial loss to add high frequency components in reconstructions. Going against the trend, we feel, the true essence of GAN lies in its ability to generate data within a completely unsupervised framework. The former method of \cite{yeh2017semantic} is thus more difficult to train because it has to `hallucinate' an entire object with just a noise/latent vector conditioning and no information of masked/damaged pixels. Thus, though, the latter school of approach has gained major attention among inpainting community, in this paper, we advocate the former genre of unsupervised approach(pixel values under mask never used). Being unsupervised is the merit of \cite{yeh2017semantic}, but it also creates a run time bottleneck. The algorithm follows iterative gradient descent optimization for finding the `best matching' noise prior corresponding to damaged image. Such iterative framework prohibits real time applications. 
 \begin{figure*}[!t]
 \centering
 \includegraphics[scale = 0.4]{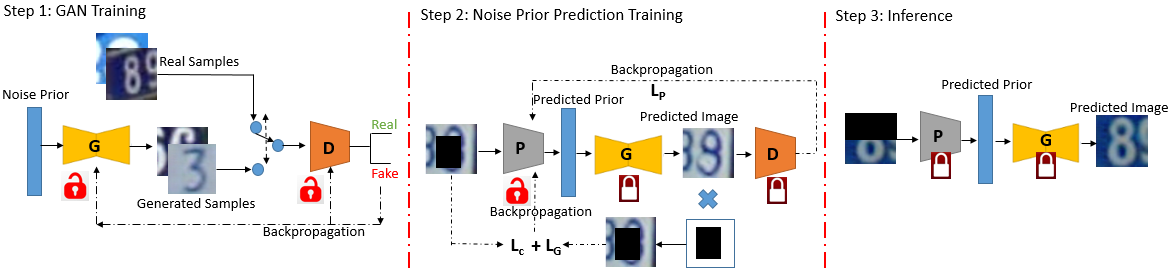}
 \caption{\scriptsize Learning and inferencing(inpainting) with our learned noise prior model. Step 1: Learn a GAN model. Step 2: Freeze GAN modules and learn to infer noise prior based on masked input image. Step 3: During inference, given a masked image, predict a matching noise vector and use pretrained GAN generator(G) to yield final output. The dashed arrows show flow of error gradients during training phase.}
 \label{fig_flow}
 \end{figure*}
\par In this paper
we primarily aim to massively accelarate inference runtime (we achieve 1500X speedup compared to \cite{yeh2017semantic}) with simultaneousness visual quality improvement  by parametrically learning noise priors. Another issue with inpainting(both supervised and unsupervised) is multi modal completion possibility of a masked region. For example, a masked lip region of face may be completed as smiling or neutral. We show that it is possible to regularize the inpainted outputs with some structural priors. As an example, for a face, we can make use of the facial landmarks as priors. Lastly, single image inpainting models cannot be appreciable applied on videos. Though each frame might be visually pleasing, when viewed as a sequence, there are lot of jitter and flicker due to temporal inconsistency of models. We propose to subdue such inconsistencies with a recurrent net based grouped noise prior learning combined with a subsequence consistency constraint. Our contributions can be summarized as follows:
\begin{enumerate}
\item Unsupervised data driven GAN noise prior prediction framework to convert the iterative paradigm of \cite{yeh2017semantic} to a single feed forward pipeline with visually better reconstruction and simultaneous massive speedup of inference time by 1500$\times$.
\item Augmenting structural priors to improve GAN samples which eventually results in better reconstructions. Such priors also regularize GAN training to respect pose and size of objects.
\item Pioneering effort towards GAN based sequence inpainting with a recurrent neural net based grouped prior learning for better temporal consistency of reconstructed sequences compared to both supervised and unsupervised benchmarks.
\item A sub-sequence consistency loss to further improve temporal smoothness of reconstructed sequences
\item We exhaustively validate our models on CelebA, SVHN, Standford Cars, CelebaHQ image datasets and VidTIMIT video dataset. 
\end{enumerate}
\section{Related works}
Traditional image inpainting methods\cite{ballester2001filling,bertalmio2000image,efros2001image,efros1999texture} broadly worked with matching patches and diffusion of low level features from unmasked sections to the masked region. These method mainly worked on synthesis of stationary textures of background scenes where it is plausible to find a matching patch from unmasked regions. However, complex objects lack such redundancy of appearance features and thus recent methods leverage hierarchical feature learning capability of deep neural nets to learn higher order semantics of a scene. Initial deep learning based methods \cite{kohler2014mask, xu2014deep} were completely supervised and trained with conservative $\mathcal{L}_2$ reconstruction loss. With the advent of GANs, a common practice \cite{context_encoders,iizuka2017globally} has been to refine the blurry reconstructions by $\mathcal{L}_2$ loss with an adversarial loss coming from a discriminator which is also simultaneously trained to distinguish real samples from inpainted samples. Notably, the first work within this paradigm of approach was Context Encoder(CE) \cite{context_encoders} by Pathak \textit{et al.}, in which the authors tried to learn scene representation along with inpainting. Iizuka \textit{et al.} proposed `Globally and Locally Consistent Image Completion' (GLCIC) in which a inpainter/generator network is pitted against two discriminators, one for gauging realism of entire image and the other for measuring fidelity of local reconstruction of masked patch region. Recently, Yu \textit{et al.} \cite{gip} improved upon GLCIC, by incorporating contextual attention within inpainting network so that the net learns to leverage distant information from uncorrupted pixels. These methods have a common pipeline of fully supervised training stage followed by adversarial loss based refinement. Thus these methods are not fully unsupervised since paired examples(masked and unmasked) are required during training.
\par In this paper, we are advocating a fully unsupervised approach (information about the masked pixels not used anywhere in training pipeline) to inpainting pioneered by Yeh \textit{et al.} \cite{yeh2017semantic}. In \cite{yeh2017semantic}, the idea is to first train a GAN framework conditioned on only noise prior($z$) sampled from some prior known distribution. At test time, since their method is completely unsupervised, the authors used an iterative gradient descent optimization to find the `best matching' $z$ vector for the damaged image with the pre-trained generator and discriminator network of the GAN. However, this iterative optimization takes about 2.5 minutes/image and is thus not suitable for practical applications. We consider the framework of \cite{yeh2017semantic} as a baseline and seek to improve upon the inference time and reconstruction quality. In the process, we also achieve comparable performance to the contemporary hybrid trained methods. 
 \begin{figure}[!t]
 \centering
 \includegraphics[scale = 0.15]{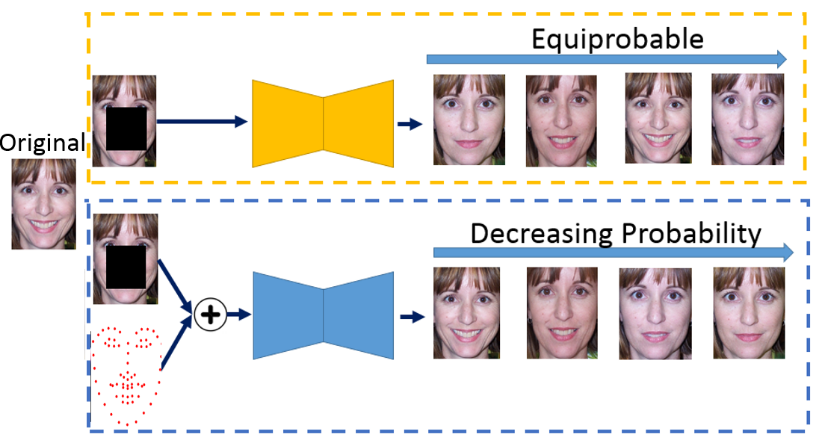}
 \caption{\scriptsize Illustration of multi model image completion possibility of  GAN based inpainting methods. Given a corrupted image, an unconditioned inpainting algorithms(top row) such as \cite{yeh2017semantic,context_encoders,iizuka2017globally,gip} samples from a uniform distribution of viable inpainted images. However, if conditioned by structural priors(bottom row), the sampling distribution is biased towards samples which preserve original facial pose and expression.}
 \label{fig_problem_multimodal}
 \end{figure}
\section{Background}
\subsection{GAN Basics}
Proposed by Goodfellow \textit{et al.}\cite{goodfellow2014generative}, a GAN model consists of two parametrized deep neural nets, viz., generator, $G_{\theta_G}$, and discriminator, $D_{\theta_D}$. The task of the generator is to yield an image, $x\in \mathcal{R}^{H\times W \times 3}$ with a latent noise prior vector, $z\in \mathcal{R}^d$, as input. $z$ is sampled from a known distribution, $p_z(z)$. A common choice \cite{goodfellow2014generative} is, $z\sim \mathcal{U}[-1,1]^d$. The discriminator is pitted against the generator to distinguish real samples(sampled from $p_{data}$) from fake/generated samples. Specifically, discriminator and generator play the following game on $V(D_{\theta_D},G_{\theta_G})$: 
$$\underset{G_{\theta_G}}{min}~~ \underset{D_{\theta_D}}{max}~~ V(D_{\theta_D}, G_{\theta_G}) = \mathbb{E}_{x\sim p_{data}(x)}[\log D_{\theta_D}(x)]$$
\vspace{-3mm}
\begin{equation}
+\mathbb{E}_{z\sim p_{z}(z)}[1 - D_{\theta_d}(G_{\theta_G}(z))].
\label{eq_gan_main_goodfellow}
\end{equation}
With enough capacity, on convergence, $G_{\theta_G}$ fools $D_{\theta_D}$ at random \cite{goodfellow2014generative}.
\subsection{Baseline GAN based unsupervised inpainting}
We first review the unsupervised inpainting baseline of Yeh \textit{et al.} \cite{yeh2017semantic}. Given a damaged image, $I_d$, corresponding to an original image, $I$, and a pre-trained GAN model, the idea is to iteratively find the `closest' $z$ vector (starting randomly from $\mathcal{U}[-1,1]^d$) which results in a reconstructed image whose semantics are similar to corrupted image. $z$ is optimized as,
\begin{equation}
\hat{z} = \underset{z}{\mathrm{argmin}}~~ L(M \odot G(z), M \odot I).
    \label{eq_yeh_loss}
\end{equation}
where $M$ is the binary mask with zeros on masked region else unity, $\odot$ is the Hadamard operator and  $L(\cdot)$ is any loss function. Interesting to note is that the loss function never makes use of pixels inside the masked region. Upon convergence, the inpainted image, $\hat{I}$, is given as, $\hat{I} = M \odot I + (1-M) \ \odot G_{\theta_G}(\hat{z})$.
\section{Proposed Method}
\subsection{Data driven Noise Prior Learning}
Though the unsupervised characteristic of \cite{yeh2017semantic} is encouraging for the generative learning community, the iterative optimization is a major bottleneck in the pipeline. Instead of iteratively optimizing the noise prior, $z$, for each test image during runtime, we propose to learn an unsupervised offline parametric model, $P_{\theta_z}$, for predicting $z$ vector. The parameter set, $\theta_z$, is optimized to minimize the following unsupervised losses: \\
\textbf{Contextual Loss:} This loss ensures that the predicted noise prior preserves fidelity with respect to the original unmasked regions.
\begin{equation}
L_{c} = M \odot (I - G(P_{\theta_z}(I_d))
\end{equation}
\textbf{Realism Loss:} This loss ensures that the inpainted output lies near the original/real data manifold and is measured by the log likelihood of belongingness to real class assigned by the pre-trained discriminator
\begin{equation}
L_{r} = \log(1 - D_{\theta_d}(G_{\theta_G}(P_{\theta_z}(I_d)))
\end{equation}
\textbf{Gradient Difference Loss:} Inspired by \cite{mathieu2015deep, nie2017medical} we also use the gradient difference loss imposed between the gradient (horizontal and vertical) matrices of original and reconstructed outputs. This compels the network to predict noise priors which yield high frequency retaining samples and also respects the gradients of the original scene.
$$L_{g} = M \odot |\nabla_x I_{d} - \nabla_x G()| $$
\begin{equation}
+ M \odot |\nabla_y I_{d} - \nabla_y G(P_{\theta_z}(I_d))|
\end{equation}
Please note that the loss is still calculated on the unmasked regions only. In summary, parameter set, $\theta_z$, is optimized to minimize the combined loss, $L_z^{com}$,
\begin{equation}
L_z^{com} = \lambda_1 L{c} + \lambda_2 L{r} + \lambda_3 L_{g},
\label{eq_total_loss}
\end{equation}
where $\lambda_i$'s controls the relative importance of each loss factor. After convergence of training of $P_{\theta_z}$, given a masked image, $I_d$, mask, $M$, we can get the inpainted output, $\hat{I}$, in one feed forward step instead of the iterative optimizations of \cite{yeh2017semantic}.
Inpainted image, $\hat{I}$, is given by, 
\begin{equation}
\hat{I} = M \odot I + (1-M) \ \odot G_{\theta_G}(P_{\theta_z}(I_d)).
\end{equation}
Though Eq. \ref{eq_yeh_loss} and \ref{eq_total_loss} are functionally same, prediction using a learned parametric network tends to perform better than ad hoc iterative optimization. This is because, with evolution of training, the network learns to adapt parameters to map images with closely matching appearances to similar $z$ vectors. Parameter update for a given image thus implicitly generalizes to images with similar characteristics.
\subsection{Regularization with Structural Priors}
Image inpainting intrinsically suffers from a multi modal completion problem. A given masked region has multiple plausible possibilities for completion. For example, consider Fig.\ref{fig_problem_multimodal}: for an unconstrained optimization setup, the masked region of the face can be inpainted with different facial expressions. From a single image inpainting point of view this might not be an issue. But in case of sequences, it is desirable to maintain a smooth flow of scene dynamics. A laughing face, for example, cannot suddenly be inpainted as a neutral frame. We propose to further regularize our network by augmenting structural priors. Structural priors can be any representation which captures the pose and size of the object to be inpainted and thereby compelling the network to yield outputs by respecting such priors. Such additional priors can be seen as conditional variables, $c$, to the GAN framework. Formulation of Eq. \ref{eq_gan_main_goodfellow} changes subtly to respect the joint distribution of real samples and conditional information. The modified game, $V(D_{\theta_D},G_{\theta_G})$: 
$$\underset{G_{\theta_G}}{min}~~ \underset{D_{\theta_D}}{max}~~ V(D_{\theta_D}, G_{\theta_G}) = \mathbb{E}_{x,c\sim p_{data}(x,c)}[\log D(x,c)]$$
\vspace{-3mm}
\begin{equation}
+\mathbb{E}_{z\sim p_{z}(z)}[1 - D_{\theta_d}(G_{\theta_G}(z))].
\label{eq_conditional_gan}
\end{equation}
The noise prior predictor network, $P_{\theta_z}$ has to optimize $\theta_z$ by respecting the structural prior as an additional constraint.
\par In this paper, without any loss of generalization, we have considered face inpainting with semantic priors as facial landmarks automatically extracted in real time(5ms @ 256$\times$256 resolution) using the robust framework of Kazemi \textit{et al.} \cite{kazemi2014one} which achieves benchmark performance on face alignment.
\subsection{Grouped Noise Prior Learning for Sequences}
\label{sec_lstm}
To our best knowledge, this is the first demonstration of GAN based completely unsupervised sequence inpainting.
A naive approach of applying the formulation of Eq. \ref{eq_total_loss} on sequences is to inpaint individual frames independently. However, such anapproach fails to learn the temporal dynamics of sequence and thereby yielding jittering effects. 
In this regard, for a sequence of $N$ frames, we propose to use a Recurrent Neural Network (RNN) to jointly predict $z$ vectors for a subset of $W$ frames at a time. RNN consist of a hidden state $h_t$ to summarize information observed upto that time step. The hidden state is updated after looking at the previous hidden state and the corrupted image(with an additional option to condition on structural priors), leading to more consistent reconstructions in terms of appearance. ames
\par
\begin{figure}[!t]
 \centering
 \includegraphics[scale = 0.2]{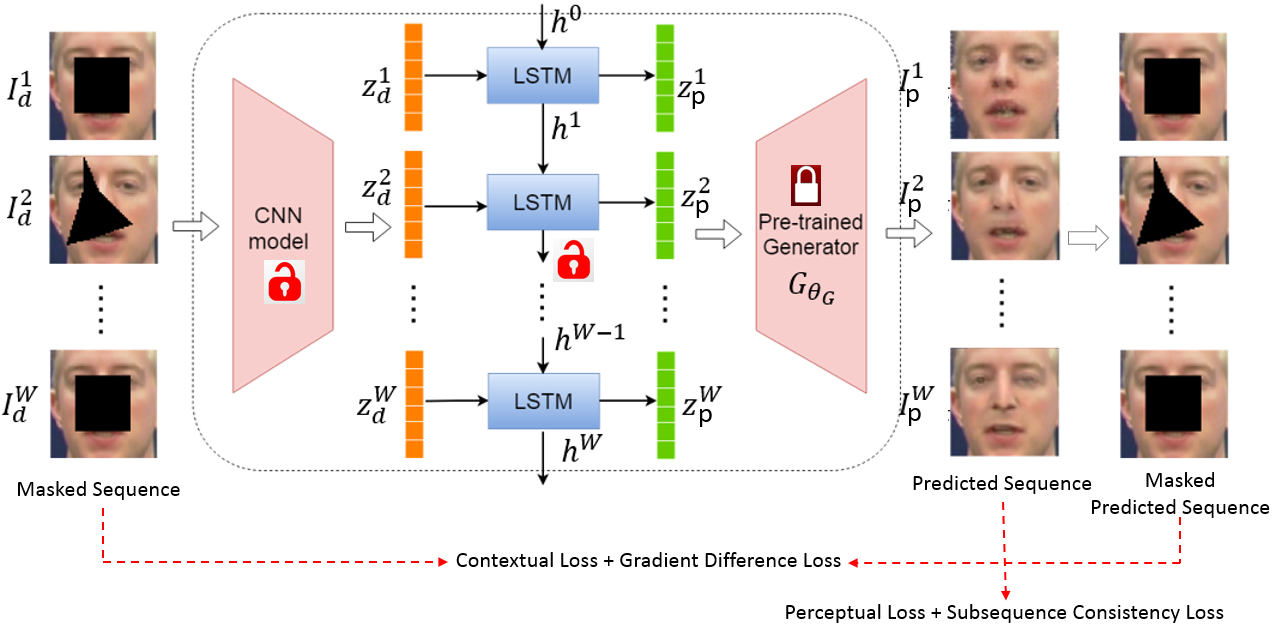}
 \caption{\scriptsize Grouped noise prior learning with a combined LSTM-CNN framework. Unlock sign means parameters to update.}
 \label{fig_lstm}
 \end{figure}

 	
Since, RNNs suffer from vanishing gradients problem\cite{bengio1994learning} and are unable to capture long dependencies, we use Long Short Term Memory (LSTM) \cite{hochreiter1997long} Networks. 
Fig. \ref{fig_lstm} shows our LSTM based framework architecture for jointly inpainting a group of frames. Let, $V=\{I_d^1,I_d^2,...,I_d^W\}$ be a group of $W$ corrupted successive frames. Initially, each frame is passed through a CNN module (same architecture of $P_{\theta_z}$, except the last layer outputs $\mathbb{R}^{d\times 1}$ instead of $\mathbb{R}^{1\times 1}$ by $P_{\theta_z}$), to obtain the input sequence for the recurrent network $z_d^k$. We obtain the predicted prior, $z_{p}^k$, by feeding the hidden state, $\{h^k\}$, of the recurrent network to a fully-connected layer.  $z_{p}^k$ is then used for reconstructions, $I_p^k$, with the help of the pre-trained generator, $G_{\theta_G}$. We use the loss function in Eq. \ref{eq_total_loss}, averaged over the grouped window of $W$ frames to optimize the parameters of LSTM and the CNN descriptor network. Specifically, the grouped prior loss is defined by, $L_z^{gr}(\cdot)$,
\begin{equation}
L_z^{gr} = \frac{1}{W} \sum_{i=1}^W L_z^{com}(I_d^i, I_p^i).
\end{equation}
Please note, the parameters of pre-trained generator and discriminator are kept frozen.  .

\subsection{Subsequence consistency loss}
We further regularize training of the LSTM framework by an implicit subsequence consistency loss over a group of $W$ neighborhood frames. The motivation is that a group of adjacent frames in a video exhibit close coherence of appearance. Thus, we define a subsequence clique as a collection of $W$ adjacent frames and penalize if the appearances of the frames differ from each other. Disparity between two inpainted images, $I_p^i$ and $I_p^j$ can be approximated by Euclidean distance between their latent vectors ($z_p^i, z_p^j$). We define the loss, $L_z^{ss}(\cdot)$ as,
\begin{equation}
L_z^{ss} = \frac{1}{\binom{W}{2}} \sum_{\forall(i,j)\in W} ||z_p^i - z_p^j||^2.
\label{eq_subsequence_loss}
\end{equation}
So, $L_z^{gr}$ helps in learning the temporal dynamics while $L_z^{ss}$ explicitly fosters temporal smoothness. If $L_z^{ss}$ dominates then, the network will be penalized by $L_z^{gr}$ because over smoothing of a sequence is not a true characterization of a real world sequence. The final loss function for the LSTM-CNN combined framework is given by $L_z^{LSTM}$,
\begin{equation}
L_z^{LSTM} = L_z^{gr} + \lambda_4L_z^{ss},
\label{eq_grouped_prior_loss}
\end{equation}
where $\lambda_4$ sets relative importance of subsequence consistency. Please note, $L_z^{ss}(\cdot)$ is applied only on a neighborhood of $W$ frames and not on entire sequence. Applying $L_z^{ss}(\cdot)$ on entire sequence is not a true representation of temporal dynamics because we will be then penalizing appearance changes even over distant frames. On contrary, reducing $W=1$ means no explicit temporal consistency loss.
 \begin{figure}[!t]
 \centering
 \includegraphics[scale = 0.45]{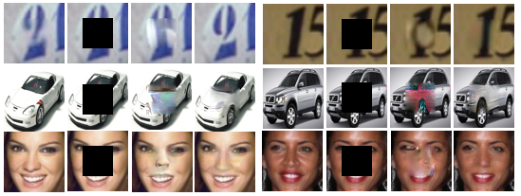}
 \caption{\scriptsize Visualization of initial inpainting solutions by the iterative framework (requires 1.5K in total) of \cite{yeh2017semantic} compared to our one feed forward pass network. Column 1: Original image; Column 2: masked image; Column 3: Initial solution by \cite{yeh2017semantic}; Column 4: \textbf{Proposed} one feed forward solution.}
 \label{fig_iterative_vs_one_shot}
 \end{figure}
 \begin{figure}[!t]
 \centering
 \includegraphics[scale = 0.28]{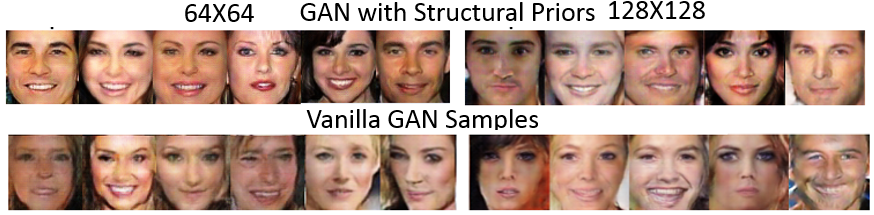}
 \caption{\scriptsize Proposed structural priors in GAN help in generating better samples compared to vanilla GAN. Random samples from proposed framework are structurally more consistent and complete. This is eventually important for better inpainting.}
 \label{fig_gan_samples} 
 \end{figure}
\section{Results}
\subsection{Single Image Inpainting}
We experiment on cropped SVHN\cite{karras2017progressive}, Standford Cars\cite{krause20133d}, CelebA\cite{liu2015deep} and CelebA-HQ\cite{liu2015deep}. SVHN crops are resized to 64$\times$64. On Standford Cars we use bounding box information to extract and resize cars to 64$\times$64. CelebA images are center cropped to 64$\times$64 and 128$\times$128. Celeb-HQ images are resized to 256$\times$256. On SVHN and Cars, we use the dataset provider's test/train split. For CelebA and CelebA-HQ, we keep 2000 images for testing.
\subsubsection{Importance of Learned Noise Prior:} The most important improvement that we achieve over \cite{yeh2017semantic} is a significant speedup during inferencing. In Fig. \ref{fig_iterative_vs_one_shot} we compare the initial solution of \cite{yeh2017semantic} with our one shot feed forward solution. Without any mechanism to estimate noise prior from masked image, initial solutions of \cite{yeh2017semantic} lie far from real data manifold and thereby mandating an iterative approach. Abiding by the suggestions in \cite{yeh2017semantic}, each image requires 1500 test time iterations. Our approach adds just subtle amount of computation for the noise predictor network and a negligible overhead for the structural priors; thereby making our model almost 1500X faster compared to \cite{yeh2017semantic}. From Fig. \ref{fig_baseline_without_mse}, it is encouraging to see that even after the iterative optimization, visual quality of our method is usually superior than \cite{yeh2017semantic}.
 \begin{figure}[!t]
 \centering
 \includegraphics[scale = 0.35]{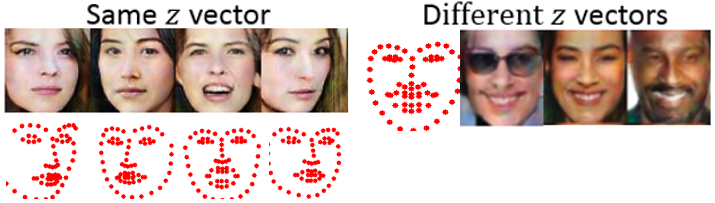}
 \caption{\scriptsize Structural priors enables GAN to disentangle facial pose and appearance cues. \textbf{Left:} Faces samples with same $z$ vector but different structural priors. \textbf{Right:} Faces sampled with different $z$ vectors for a given structural prior. Even if some keypoints are missing/occluded our model generates plausible textures.}
 \label{fig_pose_independence}
 \end{figure}
\subsubsection{Importance of Structural Priors:} 
\begin{figure}[!t]
 \centering
 \includegraphics[scale = 0.45]{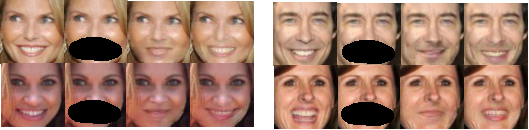}
 \caption{\scriptsize Benefit of structural prior augmented  GAN based inpainting. In each sub figure, Column 1: Original image, Column 2: Masked image, Colimn 3: Inpainted by a GAN model conditioned on proposed learned noise prior  Column 4:Inpainted by a GAN model conditioned on proposed learned noise + structural prior. Structural priors regularizes network to respect facial expression during reconstruction.}
 \label{fig_noise_vs_maps}
 \end{figure}

In this paper, we have considered the special case of face inpainting with semantic priors as facial landmarks detected by the robust framework of \cite{kazemi2014one}. We observed three fold benefits of leveraging such priors.\\
\textbf{Improved GAN Samples and Reconstructions:} Conditioning on structural priors forces the generator to yield samples closer to natural data manifold. Random samples from such conditioned generator are thus more photo-realistic (see Fig. \ref{fig_gan_samples}) compared to the unconditioned vanilla version of GAN used by \cite{yeh2017semantic}. Towards this end, we visually compare (following the protocol in \cite{shrivastava2016learning}) the  quality of random samples from our proposed semantic conditioned GAN and \cite{yeh2017semantic} at resolutions of 64$\times$64 and 128$\times$128. For visual turing test, a human annotator is randomly shown total 200 images(100 real and 100 generated) in groups of 20 and asked to label each sample as real or fake. Decisions from 10 annotators are taken. On average, 64$\times64$ resolution, the classification accuracy is 5.8\% higher for DIP($p=10^{-3}$) and 4.2\% higher($p=10^{-2}$) at 128$\times$128 resolution. Thus, human annotators found it more difficult to distinguish samples from our model compared to DIP.\\
\textbf{Control of Pose and Expression:} With structural priors, the generator learns to disentangle appearance and pose. A given semantic prior should force the generator to create a face with matching head pose and facial expression while two nearby $z$ vectors results in similar facial textures. In Figure \ref{fig_pose_independence}(Top setting) we show such disentanglement learned by our model.\\ 
\textbf{Greater structural fidelity to reference image:} In Fig. \ref{fig_noise_vs_maps}, we show the importance of structural priors on top of learned noise priors. Reconstructions with only our proposed learned noise priors might be stand alone realistic but are not penalized for changing facial expressions. For example, a (masked) smiling face can be inpainted as a neutral face by only conditioning on a learned noise prior. However, if we constrain the model with structural priors, the reconstructions are more coherent in appearance and expression to the reference image. Such structural fidelity is key in achieving temporally more consistent sequence reconstructions as discussed in upcoming sections.\\
 \begin{figure}[!t]
 \centering
 \includegraphics[scale = 0.45]{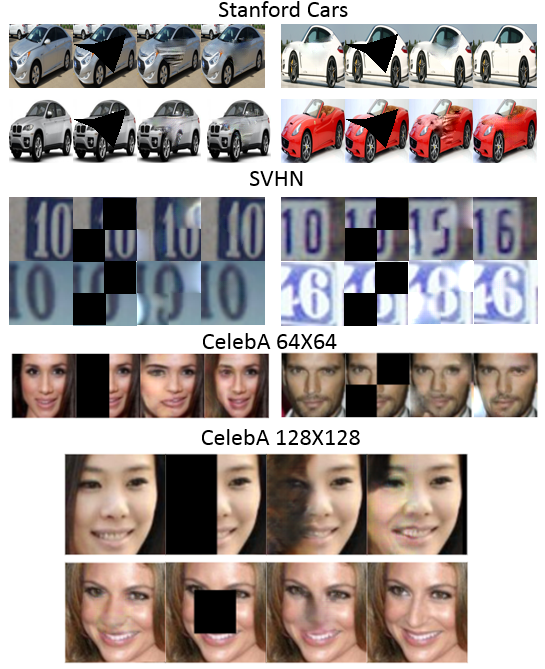}
 \caption{\scriptsize Comparative visualization of inpainting. In each set, column 1: original image, column 2: masked images, column 3: unsupervised baseline of \cite{yeh2017semantic}, column 4: \textbf{Proposed} learned noise prior conditioned model(Eq. \ref{eq_total_loss}). Proposed reconstructions are usually better, yet our model is about 1500$\times$ faster than \cite{yeh2017semantic}.}
 \label{fig_compare_z_init}
 \end{figure}
 \begin{figure*}[!t]
 \centering
 \includegraphics[scale = 0.5]{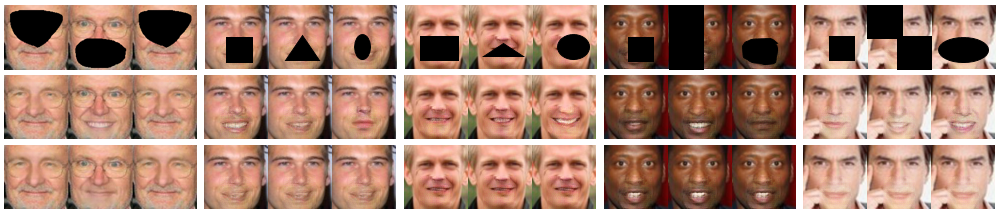}
 \caption{\scriptsize  Visualizing consistency of inpainting synthetic sequences. A synthetic sequence is created by masking a given image with different corruption patterns. Ideally we want an inpainter to yield exactly same outputs for a given synthetic sequence.; Top: Masked synthetic sequence. Middle: Inpainted sequence with Yeh \textit{et al.} \cite{yeh2017semantic}. Bottom: \textbf{Proposed} inpainted sequence with LSTM-CNN grouped prior. Proposed method yields more consistent sequences. Note, how \cite{yeh2017semantic} changes facial expressions in each frame. Proposed framework uses context from neighboring frames to improve group wise coherence. Note how lips regions are coherent even if that region is masked in some frames.}
 \label{fig_temporal_visualization} 
 \end{figure*}
 \begin{figure}[!h]
 \centering
 \includegraphics[scale = 0.2]{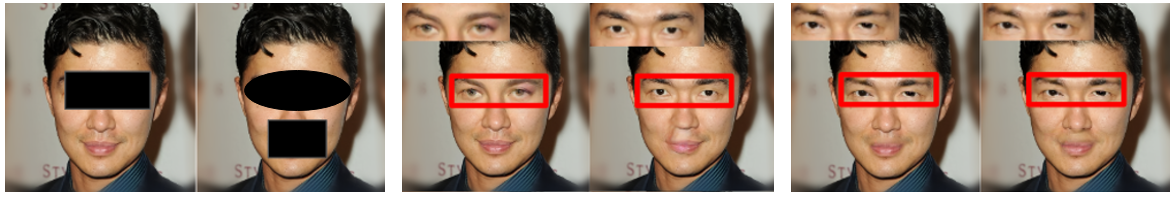}
 \caption{\scriptsize Benefit of subsequence consistency loss(Eq.\ref{eq_subsequence_loss}) augmented with grouped prior loss(Eq.\ref{eq_grouped_prior_loss}). \textbf{Left:} A synthetic sequence in which same image(sample from CelebA-HQ @ 256$\times$256) is masked differently. Ideally we want a model to inpaint both frames identically. \textbf{Middle:}Inpainting with  proposed LSTM grouped prior. \textbf{Right:} Inpainting with LSTM grouped prior + proposed subsequence consistency loss. LSTM grouped prior maintains the similarity of facial expressions(right face is inpainted neutral even though lip region was masked) but suffers from subtle texture changes(see the highlighted eye regions). Augmentation of consistency loss reduces such appearance disparities. Best viewed when zoomed up.}
 \label{fig_benefit_subsequence}
 \end{figure}
\subsubsection{Comparison to Hybrid Benchmarks}
\begin{figure}[!b]
 \centering
 \includegraphics[scale = 0.9]{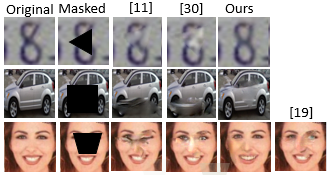}
 \caption{\scriptsize Visualization of inpainting by contemporary baselines(GLCIC\cite{iizuka2017globally}, GIP\cite{gip}, GFC \cite{gfc}) trained \textbf{without supervised reconstruction loss} but only with unsupervised adversarial loss. It is evident that training with only unsupervised loss diminishes the efficacy of the methods. Our unsupervised method however consistently shows appreciable performance. }
 \label{fig_baseline_without_mse}
 \end{figure}

\begin{figure}[!h]
 \centering
 \includegraphics[scale = 0.45]{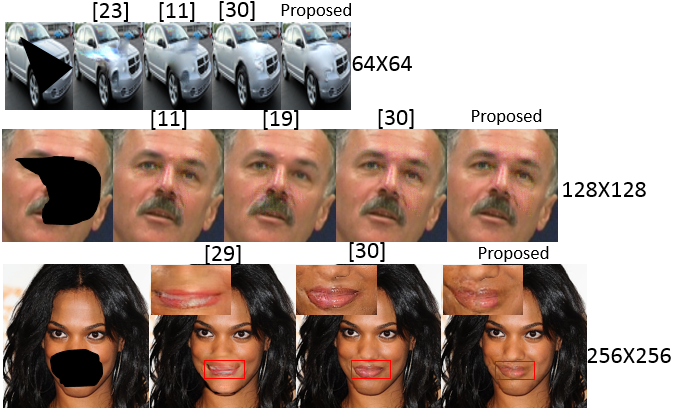}
 \caption{\scriptsize Visual comparison of inpainting with hybrid methods of CE\cite{context_encoders}, GLCIC\cite{iizuka2017globally}, GFC\cite{gfc}, GIP \cite{gip}. Our unsupervised method performs reasonably comparable even though it is totally unsupervised. Note, that at 256$\times$256(CelebA-HQ), performance of unsupervised baseline of \cite{yeh2017semantic} deteriorates drastically. Adapting the progressive training strategy in \cite{karras2017progressive} enables our GAN to mimic natural face distribution more faithfully and thereby enabling appreciable inpainting performance.}
 \label{fig_compare hybrid}
 \end{figure}

Though our method is unsupervised, for completeness of the paper, we also compare with recent hybrid inpainting benchmarks of \cite{context_encoders,iizuka2017globally, gip, gfc}. To scale up our GAN model to 256$\times$256, we follow the progressive training strategy of \cite{karras2017progressive}. See Fig. \ref{fig_compare hybrid} for visual examples. \\
\textbf{Is Supervised Phase Mandatory ?}\\
To seek an answer to this, we trained the models of \cite{context_encoders,iizuka2017globally, gip, gfc} without any $\mathcal{L}_2$ loss but only adversarial loss. We observe that these methods fail to perform in absence of $\mathcal{L}_2$ loss. In Fig. \ref{fig_baseline_without_mse}, we show some visual examples.
\subsection{Sequence Inpainting}
\subsubsection{Temporal consistency and Synthetic Sequences}
Recent deep learning based inpainting works have only been restricted for single image inpainting. The genre of video has not received interest. Even if there are some works \cite{li2018inpainting,sun2018temporally}, the reported results are in terms of per frame PSNR which does not take into account the temporal consistency/dynamics of scene reconstructions. For example, it is very annoying for a viewer if the stationary portions of a series of frames are reconstructed with different appearances on each frame and thereby creating jitter effects.
\par We dedicate this section to analyze the temporal consistencies of different methods on synthetic sequences. A synthetic sequence, $S_T$, of length $T$ is formed by taking a single image, $I$, and masking it with $T$ different/same corruptions masks. An ideal inpainting model should be agnostic of the corruption masks and yield identical reconstructions for all the $n$ frames. We define temporal consistency, $\eta_{temp}$, as the mean pairwise PSNR between all possible pairs($\hat{I}^i, \hat{I}^j$) of inpainted frames within  a synthetic sequence, $S_N$, of length, $N$;
\begin{equation} 
\eta_{temp} = \frac{1}{\binom{N}{2}} \sum_{\forall(i,j)\in S_N} PSNR(\hat{I}^i, \hat{I}^j)
\label{eq_consistency}
\end{equation}
Eq. \ref{eq_consistency} allows enumerating the consistency of a generative model. Ideally, we want $\eta_{temp}$=0. Please note that this evaluation is not possible on real videos because the transformation from one frame to another is not known and thus it is not possible to align the frames to a single frame of reference without incorporating interpolation noise with motion compensator\cite{caballero2016real}. In Table \ref{tab_consistency} we compare the consistencies with contemporary benchmarks. We see progressive improvement of consistency with the addition of LSTM-grouped prior and structural priors. Note, even the hybrid (supervised + adversarial) benchmarks manifests higher inconsistencies with exception of \cite{gfc} because it jointly trains the network with inpainting loss and face parsing loss. This bolsters the hypothesis that a prior knowledge of object structure helps in inpainting. 
\begin{table*}[!t]
\scriptsize
\centering
\caption{\scriptsize Comparison of temporal consistency in dB(Eq.\ref{eq_consistency})(left section) on synthetic sequences (2000 for CelebA, 4500 for SVHN) by competing algorithms. On right section we also report the mean PSNR of inpainting on different datasets.  Higher values of consistency are better. We compare five cases of our proposed framework; M1: Independent Learned noise($z$) prior, M2: LSTM-CNN grouped learned noise prior, M3: Independent Structural prior, M4: LSTM-CNN grouped noise prior + Structural prior, M5: LSTM-CNN grouped noise prior + Structural prior + Subsequence Consistency Loss. Masks used are RC(Random Central): random 50-70\% center mask, RF(Random Freehand): random 50\% mask by freehand mask and RCh(Random Checkboard): 50\% masked by random checkboard grid masks. \textbf{In summary}: Our unsupervised models have, in general, better temporal consistency and comparable PSNR compared to hybrid(supervised + adversarial) benchmark models.} 
\begin{tabular}{ll|l @{\hspace{1.6\tabcolsep}} l @{\hspace{1.6\tabcolsep}} l|l @{\hspace{1.6\tabcolsep}} l @{\hspace{1.6\tabcolsep}} l|l @{\hspace{1.6\tabcolsep}} l @{\hspace{1.6\tabcolsep}} l|l @{\hspace{1.6\tabcolsep}} l @{\hspace{1.6\tabcolsep}} l|l @{\hspace{1.6\tabcolsep}} l @{\hspace{1.6\tabcolsep}} l|l @{\hspace{1.6\tabcolsep}} l @{\hspace{1.6\tabcolsep}} l}\\ \hline

\multicolumn{2}{c}{} & \multicolumn{9}{c}{\hspace{-35mm} \textbf{Temporal Consistency (dB) on Synthetic Sequences} } & \multicolumn{9}{c}{\hspace{-15mm}\textbf{PSNR (dB) on Single Images}} \\    \hline 

\multicolumn{2}{c}{} & \multicolumn{3}{c}{SVHN @ 64} & \multicolumn{3}{c}{CelebA @ 128} & \multicolumn{3}{c}{Cars@ 64} & \multicolumn{3}{c}{SVHN@ 64} & \multicolumn{3}{c}{CelebA@ 128} & \multicolumn{3}{c}{CelebA-HQ @ 256} \\    \hline   

 Genre   & Method & RC & RF & RCh & RC     & RF   & RCh & RC     & RF   & RCh & RC     & RF   & RCh & RC     & RF   & RCh & RC     & RF   & RCh \\\hline
                                                
& CE\cite{context_encoders}  & 20.5     & 22.0     & 21.3    & 20.1     & 21.4     & 19.8 & 14.2 & 14.8 & 13.9 & 21.9 & 22.0 & 21.8 & 24.3 & 25.0 &24.0  & 17.8& 18.2& 17.7  \\
                                                                  \begin{tabular}[c]{@{}l@{}}Sup.+\end{tabular} & GLCIC\cite{iizuka2017globally}  & 21.6   & 22.0  & 21.7 & 20.9   & 22.1  & 20.1 & 15.9& 16.9& 15.7& 23.2& 24.0& 22.7& 27.9& 28.0& 27.2& 23.8& 23.7& 22.7 \\
\begin{tabular}[c]{@{}l@{}}Adv\end{tabular} & GIP\cite{gip}                                                                                                          & 21.7 & 22.9 & 23.0  & 21.1 & 22.2 & 21.3 & 16.1 & 17.2 & 15.9 & 23.9 & 24.1&23.0& 28.2 & 28.7& 27.7 & 24.1 & 24.3&23.6 \\
& GFC \cite{gfc}  & - & - & - & 23.1   & 24.9  & 23.8 &-&-&-&-&-&- &28.0 & 28.2 & 27.1 & - & - & - \\
\hline
                                                                  & Yeh \textit{et al.}\cite{yeh2017semantic}  & 22.5     & 22.9    & 22.8    & 21.9    & 22.2    & 21.1 & 13.9 & 14.0& 13.2& 20.9 & 21.2& 21.0 & 23.0 & 23.1 & 21.4& 15.7 & 16.0 & 13.1  \\
                                                                  & \begin{tabular}[c]{@{}c@{}}Proposed:M1 \end{tabular}                                   & 23.8     & 25.9     & 24.2   & 22.6     & 24.0     & 23.0 & 15.2 & 15.6 & 15.1 & 23.0 & 23.8 & 22.5 & 24.8& 25.2 & 23.7 & 20.1 & 20.4& 18.9  \\
Unsup                                                      & \begin{tabular}[c]{@{}c@{}}Proposed:M2 \end{tabular}                                      & 25.0     & 26.9     & 25.9   & 23.8     & 25.6     & 24.2 & -&-&-&-&-&-&-&-&-&-&-&-    \\

& \begin{tabular}[c]{@{}c@{}}Proposed:M3\end{tabular} & - & - & - & 24.1     & 27.4     & 27.1 &-&-&-&-&-&-& 27.4& 27.9 &26.4 & 22.6& 23.0 & 22.0 \\

& \begin{tabular}[c]{@{}c@{}}Proposed:M4\end{tabular} & - & - & - & 26.3     & 29.8     & 29.4 &-&-&-&-&-&-&-&-&-&-&-&- \\

& \begin{tabular}[c]{@{}c@{}}Proposed:M5 \end{tabular}                   & -    & -     & -   & 27.6 & 28.0 & 26.9  &-&-&-&-&-&-&-&-&-&-&-&-  \\

\hline
\end{tabular}
\label{tab_consistency}
\end{table*} 
\begin{figure*}[t]
\centering
   \begin{subfigure}{0.49\linewidth} \centering
     \includegraphics[scale = 0.25]{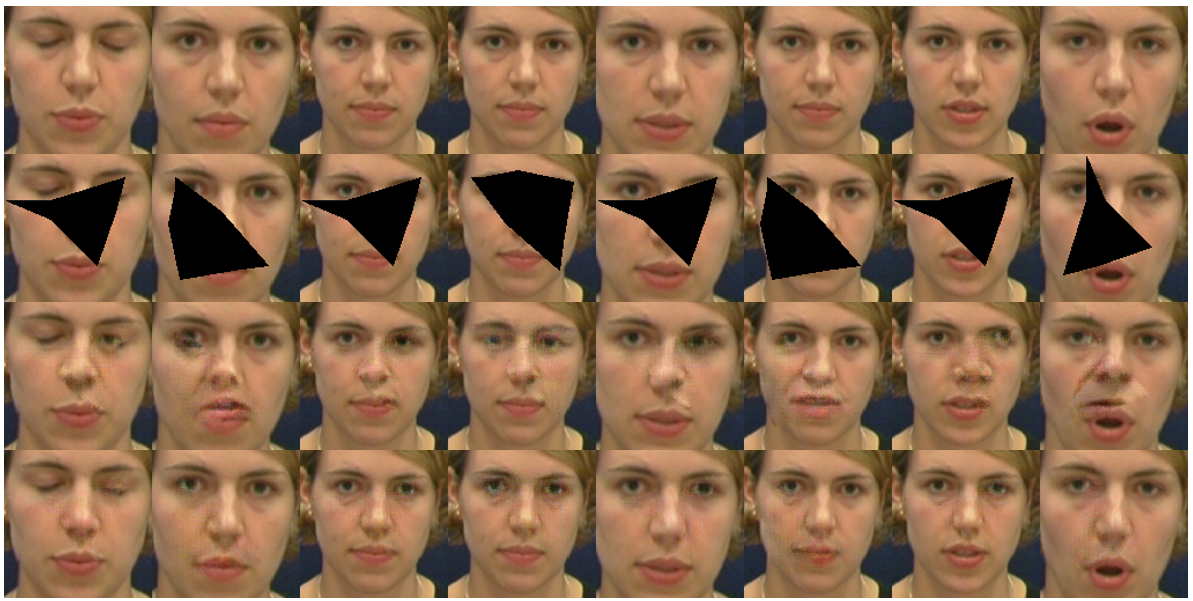}
   \end{subfigure}
   \begin{subfigure}{0.49\linewidth} \centering
     \includegraphics[scale = 0.25]{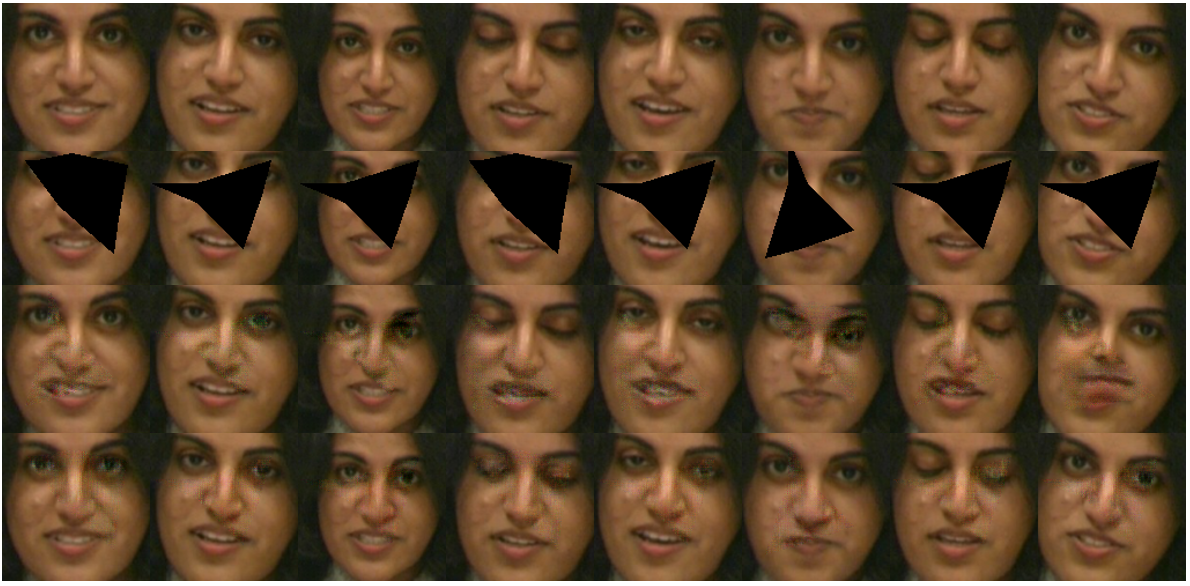}
   \end{subfigure}\\
   \caption{\scriptsize Inpainting on VidTIMIT sequences by \cite{yeh2017semantic}(3$^{rd}$ row) and proposed method: version M5(4$^{th}$ row). See Table\ref{tab_consistency} for definition of M5. }
   \label{fig_gpu_run_128_glcic}
   \end{figure*}
\subsubsection{Importance of Subsequence Consistency Loss}
In Fig. \ref{fig_benefit_subsequence} we show a synthetic sequence, in which a same face is masked differently. Proposed LSTM-grouped prior based reconstruction is successful is maintaining the overall same facial expression but fails to maintain subtle textural consistencies as shown in highlighted insets. Subsequence consistency loss helps in maintaining such subtle texture coherence which results in improved temporal consistency. Again, please note, these difference are much more easier to illustrate(and visualize) in such synthetic sequences than on real videos.
\subsubsection{Application on Real Videos:}
The experiments with synthetic sequences taught us three lessons, viz., a)LSTM-CNN based grouped noise prior learning is better than independent noise prior learning b) structural prior fosters in higher fidelity and c) subsequence consistency loss helps in preserving subtle texture details. With these knowledge, we proceed to demonstrate first attempt towards GAN based inpainting on real videos. For this, we selected the VidTIMIT dataset\cite{sanderson2009multi} which consists of video recordings of 43 subjects each narrating 10 different sentences. Images of CelebA dataset are of superior resolution than those of VidTIMIT. Due to this intrinsic difference of data distribution we finetuned our pretrained(trained on CelebA) models on randomly selected 33 subjects of VidTIMIT.  Remaining videos of 10 subjects were kept for testing inpainting performances. In total, there are total 9600 frames for testing. All faces center cropped to 128$\times$128.\\ 
\textbf{Evaluating Video Quality: MOVIE metric \cite{movie}:} Traditional metrics such as PSNR and structure similarity (SSIM) are not a true reflection of human visual perception measure as shown in recent studies \cite{gip,srgan}. Also, these metrics donot consider any temporal information. For this, we preferred to use the MOVIE metric \cite{movie}. MOVIE is a spatio-spectrally localized framework for assessing a video quality by considering spatial, temporal and spatio-temporal aspects. A lower value of MOVIE metric indicates a better video. MOVIE metric was found to appreciably correlate with human perception. In Table \ref{tab_movie_metric}, we compare the average test set MOVIE metric. All variants of our proposed framework outperforms \cite{yeh2017semantic}. With independent noise prior model, we get better performance than \cite{context_encoders} and comparable performance to \cite{iizuka2017globally,gip}. Addition of LSTM-grouped prior and structural prior boosts our performance with further improvement coming from subsequence consistency loss. It is interesting to see that even if we compute structural prior on third(and reuse in between), there is subtle degradation of performance. We show some video snippets in Fig. \ref{fig_gpu_run_128_glcic}.
\begin{table}[!h]
\scriptsize
\caption{\scriptsize Comparison of MOVIE metric\cite{movie} averaged over test sequences of VidTIMIT dataset. Lower value of metric is better for perceptual quality of a reconstructed sequence. Refer to Table \ref{tab_consistency} for definition of proposed methods M1-M5. M6 is framework of M5 but structural priors are evaluated every alternate third frame.}
\label{my-label}
\begin{tabular}{cccccccccc}\\\hline
\multicolumn{5}{c}{Competing} & \multicolumn{5}{c}{Proposed} \\\hline
\cite{yeh2017semantic} & \cite{context_encoders} & \cite{iizuka2017globally} & \cite{gip} & \cite{gfc} & M1 & M2 & M4 & M5 & M6 \\
0.68 & 0.60 & 0.52 & 0.42 & 0.35 & 0.48 & 0.31 & 0.23 & 0.18 & 0.22 \\\hline
\end{tabular}
\label{tab_movie_metric}
\end{table}
\section{Discussion and Conclusion}
In this paper, we showed the importance of priors in GANs for pushing the performance envelope of unsupervised inpainting framework of \cite{yeh2017semantic} with better inpainting quality and almost 1500$\times$ speedup. The objective of this paper was to purposefully abstain from the contemporary practice of hybrid(supervised + unsupervised) training and focus on creating a faster unsupervised framework comparable visual performance. Our proposed framework with grouped LSTM-CNN guided noise prior and structural  prior manifests better spatio-temporal characteristics than contemporary hybrid baselines. This shows that current single image inpainting methods have further scopes of improvement on videos and the frameworks used by us in this regard can be exploited by those algorithms also. Given the current state of GAN research, it is not expected that a completely unsupervised GAN based inpainter can work on natural images such as ImageNet or Places2 dataset(which hybrid methods are capable of due to supervised $\mathcal{L}_2$). However, as our understanding on GANs improve and we enable GAN models to generate natural scenes, the methods of this paper shall seamlessly fit in those scenarios as well.
{\small
\bibliographystyle{ieee}
\bibliography{egbib}

\begin{thebibliography}{10}\itemsep=-1pt

\bibitem{ballester2001filling}
C.~Ballester, M.~Bertalmio, V.~Caselles, G.~Sapiro, and J.~Verdera.
\newblock Filling-in by joint interpolation of vector fields and gray levels.
\newblock {\em IEEE transactions on image processing}, 10(8):1200--1211, 2001.

\bibitem{barnes2009patchmatch}
C.~Barnes, E.~Shechtman, A.~Finkelstein, and D.~B. Goldman.
\newblock Patchmatch: A randomized correspondence algorithm for structural
  image editing.
\newblock {\em ACM Transactions on Graphics (ToG)}, 28(3):24, 2009.

\bibitem{bengio1994learning}
Y.~Bengio, P.~Simard, and P.~Frasconi.
\newblock Learning long-term dependencies with gradient descent is difficult.
\newblock {\em IEEE transactions on neural networks}, 5(2):157--166, 1994.

\bibitem{bertalmio2000image}
M.~Bertalmio, G.~Sapiro, V.~Caselles, and C.~Ballester.
\newblock Image inpainting.
\newblock In {\em Proceedings of the 27th annual conference on Computer
  graphics and interactive techniques}, pages 417--424. ACM
  Press/Addison-Wesley Publishing Co., 2000.

\bibitem{caballero2016real}
J.~Caballero, C.~Ledig, A.~Aitken, A.~Acosta, J.~Totz, Z.~Wang, and W.~Shi.
\newblock Real-time video super-resolution with spatio-temporal networks and
  motion compensation.
\newblock {\em CVPR}, 2016.

\bibitem{efros2001image}
A.~A. Efros and W.~T. Freeman.
\newblock Image quilting for texture synthesis and transfer.
\newblock In {\em Proceedings of the 28th annual conference on Computer
  graphics and interactive techniques}, pages 341--346. ACM, 2001.

\bibitem{efros1999texture}
A.~A. Efros and T.~K. Leung.
\newblock Texture synthesis by non-parametric sampling.
\newblock In {\em ICCV}, page 1033. IEEE, 1999.

\bibitem{goodfellow2014generative}
I.~Goodfellow, J.~Pouget-Abadie, M.~Mirza, B.~Xu, D.~Warde-Farley, S.~Ozair,
  A.~Courville, and Y.~Bengio.
\newblock Generative adversarial nets.
\newblock In {\em NIPS}, pages 2672--2680, 2014.

\bibitem{hays2007scene}
J.~Hays and A.~A. Efros.
\newblock Scene completion using millions of photographs.
\newblock In {\em ACM Transactions on Graphics (TOG)}, volume~26, page~4. ACM,
  2007.

\bibitem{hochreiter1997long}
S.~Hochreiter and J.~Schmidhuber.
\newblock Long short-term memory.
\newblock {\em Neural computation}, 9(8):1735--1780, 1997.

\bibitem{iizuka2017globally}
S.~Iizuka, E.~Simo-Serra, and H.~Ishikawa.
\newblock Globally and locally consistent image completion.
\newblock {\em ACM Transactions on Graphics (TOG)}, 36(4):107, 2017.

\bibitem{karras2017progressive}
T.~Karras, T.~Aila, S.~Laine, and J.~Lehtinen.
\newblock Progressive growing of gans for improved quality, stability, and
  variation.
\newblock In {\em ICLR}, 2018.

\bibitem{kazemi2014one}
V.~Kazemi and J.~Sullivan.
\newblock One millisecond face alignment with an ensemble of regression trees.
\newblock In {\em CVPR}, pages 1867--1874, 2014.

\bibitem{kingma2013auto}
D.~P. Kingma and M.~Welling.
\newblock Auto-encoding variational bayes.
\newblock {\em arXiv preprint arXiv:1312.6114}, 2013.

\bibitem{kohler2014mask}
R.~K{\"o}hler, C.~Schuler, B.~Sch{\"o}lkopf, and S.~Harmeling.
\newblock Mask-specific inpainting with deep neural networks.
\newblock In {\em German Conference on Pattern Recognition}, pages 523--534.
  Springer, 2014.

\bibitem{krause20133d}
J.~Krause, M.~Stark, J.~Deng, and L.~Fei-Fei.
\newblock 3d object representations for fine-grained categorization.
\newblock In {\em Proceedings of the IEEE International Conference on Computer
  Vision Workshops}, pages 554--561, 2013.

\bibitem{srgan}
C.~Ledig, L.~Theis, F.~Husz{\'a}r, J.~Caballero, A.~Cunningham, A.~Acosta,
  A.~P. Aitken, A.~Tejani, J.~Totz, Z.~Wang, et~al.
\newblock Photo-realistic single image super-resolution using a generative
  adversarial network.
\newblock In {\em CVPR}, volume~2, page~4, 2017.

\bibitem{li2018inpainting}
C.~Li, Y.~Ding, B.~Yu, M.~Xu, and Q.~Zhang.
\newblock Inpainting of continuous frames of old movies based on deep neural
  network.
\newblock In {\em 2018 International Conference on Audio, Language and Image
  Processing (ICALIP)}, pages 132--137. IEEE, 2018.

\bibitem{gfc}
Y.~Li, S.~Liu, J.~Yang, and M.-H. Yang.
\newblock Generative face completion.
\newblock In {\em The IEEE Conference on Computer Vision and Pattern
  Recognition (CVPR)}, volume~1, page~3, 2017.

\bibitem{liu2015deep}
Z.~Liu, P.~Luo, X.~Wang, and X.~Tang.
\newblock Deep learning face attributes in the wild.
\newblock In {\em Proceedings of the IEEE International Conference on Computer
  Vision}, pages 3730--3738, 2015.

\bibitem{mathieu2015deep}
M.~Mathieu, C.~Couprie, and Y.~LeCun.
\newblock Deep multi-scale video prediction beyond mean square error.
\newblock {\em ICLR}, 2016.

\bibitem{nie2017medical}
D.~Nie, R.~Trullo, J.~Lian, C.~Petitjean, S.~Ruan, Q.~Wang, and D.~Shen.
\newblock Medical image synthesis with context-aware generative adversarial
  networks.
\newblock In {\em MICCAI}, pages 417--425. Springer, 2017.

\bibitem{context_encoders}
D.~Pathak, P.~Krahenbuhl, J.~Donahue, T.~Darrell, and A.~A. Efros.
\newblock Context encoders: Feature learning by inpainting.
\newblock In {\em CVPR}, pages 2536--2544, 2016.

\bibitem{sanderson2009multi}
C.~Sanderson and B.~C. Lovell.
\newblock Multi-region probabilistic histograms for robust and scalable
  identity inference.
\newblock In {\em International Conference on Biometrics}, pages 199--208.
  Springer, 2009.

\bibitem{movie}
K.~Seshadrinathan and A.~C. Bovik.
\newblock Motion tuned spatio-temporal quality assessment of natural videos.
\newblock {\em IEEE transactions on image processing}, 19(2):335--350, 2010.

\bibitem{shrivastava2016learning}
A.~Shrivastava, T.~Pfister, O.~Tuzel, J.~Susskind, W.~Wang, and R.~Webb.
\newblock Learning from simulated and unsupervised images through adversarial
  training.
\newblock {\em CVPR}, pages 2107--2116, 2017.

\bibitem{sun2018temporally}
X.~Sun, R.~Szeto, and J.~J. Corso.
\newblock A temporally-aware interpolation network for video frame inpainting.
\newblock {\em arXiv preprint arXiv:1803.07218}, 2018.

\bibitem{xu2014deep}
L.~Xu, J.~S. Ren, C.~Liu, and J.~Jia.
\newblock Deep convolutional neural network for image deconvolution.
\newblock In {\em Advances in Neural Information Processing Systems}, pages
  1790--1798, 2014.

\bibitem{yeh2017semantic}
R.~A. Yeh, C.~Chen, T.~Y. Lim, A.~G. Schwing, M.~Hasegawa-Johnson, and M.~N.
  Do.
\newblock Semantic image inpainting with deep generative models.
\newblock In {\em CVPR}, pages 5485--5493, 2017.

\bibitem{gip}
J.~Yu, Z.~Lin, J.~Yang, X.~Shen, X.~Lu, and T.~S. Huang.
\newblock Generative image inpainting with contextual attention.
\newblock In {\em CVPR}, 2018.

\end{thebibliography}
}
\end{document}